\documentclass[10pt,twocolumn,letterpaper]{article}

\usepackage{cvpr}
\usepackage{times}
\usepackage{epsfig}
\usepackage{graphicx}
\usepackage{amsmath}
\usepackage{amssymb}

\usepackage[pagebackref=true,breaklinks=true,letterpaper=true,colorlinks,bookmarks=false]{hyperref}

\cvprfinalcopy 

\def\cvprPaperID{****} 

\ifcvprfinal\fi
\begin{document}

\title{Alleviating Noisy Data in Image Captioning  with Cooperative Distillation}

\author{Pierre Dognin$^*$\and
		Igor Melnyk$^*$\and
		Youssef Mroueh$^*$\and
		Inkit Padhi$^*$\and
		Mattia Rigotti$^*$\and
		Jarret Ross$^*$\and
		 \\ IBM Reseach AI \and
		Yair Schiff$^*$\\
		$*$ Alphabetical order
}

\maketitle

\begin{abstract}
Image captioning systems have made substantial progress, largely due to the availability of curated datasets like Microsoft COCO or Vizwiz that have accurate descriptions of their corresponding images.
Unfortunately, scarce availability of such cleanly labeled data results in trained algorithms producing captions that can be terse and idiosyncratically specific to details in the image.
We propose a new technique, \emph{cooperative distillation} that combines clean curated datasets with the web-scale automatically extracted captions of the Google Conceptual Captions dataset (GCC), which can have poor descriptions of images, but is abundant in size and therefore provides a rich vocabulary resulting in more expressive captions.
\end{abstract}

Learning with noisy data is an important and challenging problem in machine learning.
In image captioning, the problem of noisy annotations has been addressed by several authors.
For example, \cite{GCC} uses a number of heuristics (tagging, annotations, word statistics, etc.) to filter out low quality captions.
Similarly, in \cite{zhang2019denoising} the training captions are denoised by extracting only the noun phrases, which are then treated as a training dataset.
In the winning entry of the GCC challenge \cite{googlewinner}, authors employ dynamic filtering in the training, which is based on model confidence.

Our cooperative distillation (co-distill) framework trains a \emph{student model} on a large \emph{noisy dataset}.
By noise we mean that captions are often grammatically incorrect or do not match the semantic content of images.
We also rely on a \emph{clean dataset} to train a \emph{teacher model}.
We explore whether we can leverage the specific advantages of both types of datasets by training on a rich vocabulary and variety of scene contexts, while alleviating the noisy annotations.


\noindent\textbf{Semantic Bridge between Datasets.}
We use a BERT tokenizer to generate a joint vocabulary across both the noisy and clean datasets at a \emph{sub-word units} level, and we also leverage BERT to represent captions as the embedding of the resulting [CLS] token.
Such representation has been shown to capture semantic information that can be used to quantify the \emph{semantic similarity} between captions (see e.g.\ \cite{reimers2019sentence}).
Semantic similarity is a crucial element in our algorithm used for: 1) \emph{denoising}, by down-weighting noisy captions that are not semantically aligned with those generated by the teacher model, and 2) \emph{increasing caption diversity}, by augmenting the training dataset with generated captions that are aligned with clean ground truth captions.

\noindent\textbf{Student and Teacher Transformer Models.}
Transformer networks from \cite{Transformer} are used for both student and teacher models.
Captions are generated conditionally on the encoded features via the transformer decoder.
The transformer architectures have $2$ layers (for both encoder and decoder), embeddings of size $512$, and $8$ attention heads per layer.
\begin{figure*}[t!]
    \centering
 	\resizebox{0.95\linewidth}{!}{%
 	    \includegraphics[width=\linewidth]{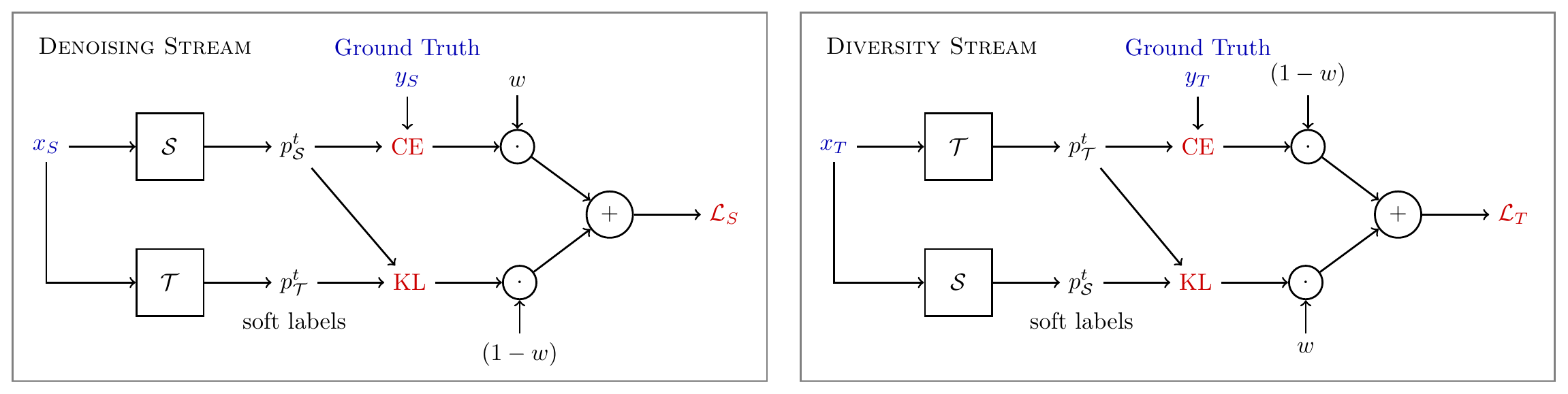}
 	    } 
 	\caption{Denoising and diversity streams.
 	    In the denoising stream (left panel), an image $x_S$ from the clean dataset $S$ is decoded by both the student model $\mathcal{S}$ and teacher model $\mathcal{T}$, resulting in soft-label predictions (softmaxes) $\{p^t_{\mathcal{S}}\}_t$ and $\{p^t_{\mathcal{T}}\}_t$.
 	    These are used to compute the Cross-Entropy (CE) loss and the KL divergence in the distillation loss, which in turn are weighted by the semantic coherence weight $w$.
 	    The diversity stream (right panel) is analogous to the denoising stream with the role of student and teacher reversed.
 	    For high similarity weights, $w$, the teacher is trained by distilling the student.
 	 }
	\label{fig:train}
 \vskip -0.1in
 \end{figure*}

\noindent \textbf{Cooperative Distillation.}
In order to train both the student and the teacher, we alternate between two training streams: denoising and diversity, shown in Figure \ref{fig:train}.
In the \emph{denoising stream}, the student minimizes its loss given a fixed teacher model, and in the \emph{diversity stream}, the teacher minimizes its loss given a fixed student as explained next.

\noindent\textbf{Denoising Stream Loss.} 
Given a sample from the noisy student dataset, we embed the ground truth caption through BERT and represent it as the embedding of the resulting [CLS] token.
We then push the image through the Teacher Transformer to obtain a softmax distribution, which is decoded using a greedy-max approach to produce the teacher's predicted tokens sequence.
As with the tokenized ground truth caption, we embed this predicted sequence via BERT.
These two embeddings allow us to define a \emph{semantic coherence weighting} for the denoising stream, which captures the similarity between the noisy ground truth caption and the caption predicted by the teacher model.

For a fixed teacher, the student minimizes a per-sample loss that combines weighted terms of cross-entropy based on ground truth labels and KL-divergence between student and teacher predictions.
The weight on each term is determined by the semantic coherence calculated through the BERT embedding bridge.
If the semantic coherence weight is high, the student can ``trust'' its ground truth and puts a higher weight on the regular cross-entropy training term.
If the semantic coherence weight is low, the student instead distills the teacher through a KL-term comparing the soft-labels from the student and the caption predicted by the teacher.
Hence, the denoising distillation loss interpolates between the hard noisy label and the teacher's soft-label.

\noindent \textbf{Diversity Stream Loss.}
For a sample from the clean teacher dataset, we follow a similar procedure.
We begin by embedding the clean ground truth caption via BERT.
We then obtain predicted captions from the student model  for each image using greedy max decoding.
We embed the student's predicted caption using BERT.
Finally, we define the semantic coherence weight between the teacher ground truth caption and the caption predicted by the student.

For a fixed student the teacher minimizes a per sample loss that combines weighted terms of cross-entropy based on ground truth labels and KL-divergence between teacher and student predictions.
The teacher trusts its ground truth caption if the semantic coherence weight is small, and distills the student if the semantic coherence weight between the ground truth caption and the caption predicted by the student is high.
This injects language diversity into the teacher model training for captions with high weight. 

\begin{figure}[ht!]
 \centering
 \includegraphics[width=0.6\linewidth]{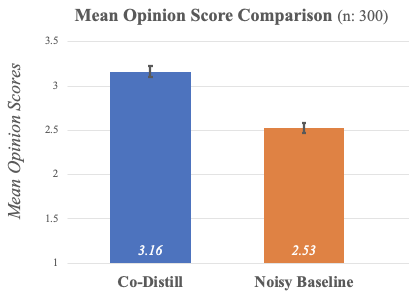}

 \caption{
 	Human evaluations of noisy baseline and co-distill.
 	}
 	\label{fig:mturk}
 \vskip -0.1in
\end{figure}

\noindent \textbf{Results.}
 We use two datasets for our experiments. The first is \emph{Microsoft COCO} \cite{MSCOCO} (the clean dataset) and a subsest of 500K images of Google \emph{Conceptual Captions} (GCC) \cite{GCC} (the noisy dataset). We conduct a human  evaluation on Amazon MTurk, where human evaluators where shown an image with captions from our co-distill method and from a baseline trained on the noisy dataset and evaluators were asked to rate each caption on a Likert scale from 1 to 5. We see from Figure \ref{fig:mturk} that co-distill outperforms the noisy baseline, hence effectively denoising the noisy training data.  
 \begin{figure}[ht!]
 \centering
 \includegraphics[width=1.0\linewidth]{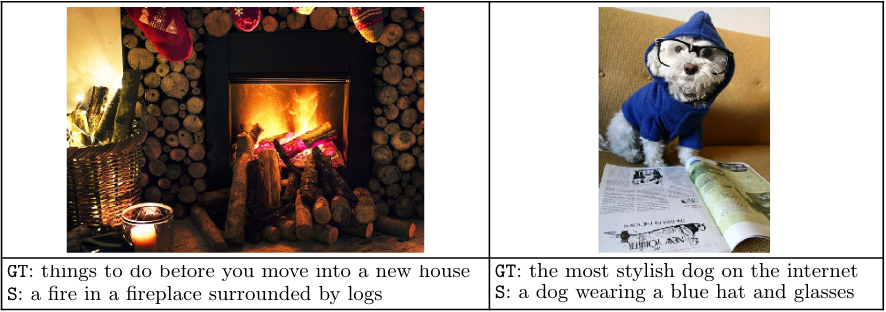}

 \caption{
 	Example captions on GCC test dataset. Captions generated by the student (S) tend to be more descriptive and image-specific than the ``noisy'' groundtruth annotations (GT).
 	}
 	\label{fig:example}
 \vskip -0.0in
\end{figure}

{\small
\bibliographystyle{ieee_fullname}
\bibliography{egbib}

\begin{thebibliography}{1}\itemsep=-1pt

\bibitem{MSCOCO}
Tsung{-}Yi Lin, Michael Maire, Serge~J. Belongie, Lubomir~D. Bourdev, Ross~B.
  Girshick, James Hays, Pietro Perona, Deva Ramanan, Piotr Doll{\'{a}}r, and
  C.~Lawrence Zitnick.
\newblock Microsoft {COCO:} common objects in context.
\newblock {\em EECV}, 2014.

\bibitem{googlewinner}
Ruotian Luo, Gilad Vered, Lior Bracha, Gal Chechik, and Greg Shakhnarovich.
\newblock {Winning Google Conceptual captions challenge}, 2019.

\bibitem{reimers2019sentence}
Nils Reimers and Iryna Gurevych.
\newblock Sentence-{BERT}: Sentence embeddings using siamese {BERT}-networks.
\newblock {\em arXiv preprint arXiv:1908.10084}, 2019.

\bibitem{GCC}
Piyush Sharma, Nan Ding, Sebastian Goodman, and Radu Soricut.
\newblock Conceptual captions: A cleaned, hypernymed, image alt-text dataset
  for automatic image captioning.
\newblock In {\em Proceedings of the 56th Annual Meeting of the Association for
  Computational Linguistics (Volume 1: Long Papers)}, pages 2556--2565, July
  2018.

\bibitem{Transformer}
Ashish Vaswani, Noam Shazeer, Niki Parmar, Jakob Uszkoreit, Llion Jones,
  Aidan~N Gomez, \L~ukasz Kaiser, and Illia Polosukhin.
\newblock Attention is all you need.
\newblock In {\em NeurIPS}, pages 5998--6008. 2017.

\bibitem{zhang2019denoising}
Yulong Zhang, Yuxin Ding, Rui Wu, and Fuxing Xue.
\newblock {A Denoising Framework for Image Caption}.
\newblock In {\em 2019 IEEE Intl Conf on Dependable, Autonomic and Secure
  Computing}, pages 825--832. IEEE, 2019.

\end{thebibliography}
}

\end{document}